%% file: main.tex
\documentclass[10pt,letterpaper,twocolumn]{article}
\usepackage[T1]{fontenc}
\usepackage[utf8]{inputenc}
\usepackage[margin=1in,columnsep=0.25in]{geometry}
\usepackage{newtxtext}
\usepackage[hyphens]{url}
\usepackage{graphicx}
\usepackage[authoryear,round]{natbib}
\usepackage[font=small,labelfont=bf]{caption}
\usepackage{booktabs,array,adjustbox}
\usepackage{algorithm,algorithmic}
\usepackage{placeins,multicol}

\input{math_commands}

\usepackage[hidelinks,unicode]{hyperref}
\makeatletter
\renewcommand\section{\@startsection{section}{1}{\z@}%
  {-2.0ex plus -0.5ex minus -0.2ex}{0.8ex plus 0.2ex}{\normalfont\large\bfseries}}
\renewcommand\subsection{\@startsection{subsection}{2}{\z@}%
  {-1.5ex plus -0.5ex minus -0.2ex}{0.6ex plus 0.2ex}{\normalfont\normalsize\bfseries}}
\makeatother

\title{\bfseries TOLA: Text-aware One-Step Latent Adaptation\\
for Diffusion-based Text Image Super-Resolution}
\input{authors}
\date{}
\hypersetup{
  pdftitle={TOLA: Text-aware One-Step Latent Adaptation for Diffusion-based Text Image Super-Resolution},
  pdfsubject={Text image super-resolution},
  pdfkeywords={text image super-resolution, diffusion, one-step adaptation}
}

\begin{document}

\maketitle

\begin{abstract}
Text image super-resolution (TSR) aims to recover visually faithful and readable text under unknown degradations. Existing diffusion-based methods typically rely on multi-step prediction of either the high-resolution image or its text prior, resulting in prohibitive computational cost and inference latency. More critically, an erroneous text prior may be repeatedly injected into the denoising process, causing image and text predictions to reinforce each other and progressively amplify an early recognition error into a sharp yet semantically incorrect character. To address these limitations, we propose TOLA, a Text-aware One-step Latent Adaptation framework without iterative image-text diffusion. TOLA consists of two key modules. First, a confidence-weighted text conditioning module constructs the semantic condition only once and suppresses unreliable OCR predictions before they contaminate image reconstruction. Second, a lightweight latent residual correction module explicitly estimates and corrects the structured residual errors to recover missing or distorted stroke details. Extensive experiments demonstrate our state-of-the-art performance across all evaluation metrics on both CTR-TSR-Test ($\times$4) and RealCE-200 benchmarks. It is worth noting that our TOLA  consistently surpasses existing diffusion-based TSR methods by at least 2.72 dB in PSNR on CTR-TSR-Test.

\end{abstract}

\section{Introduction}
Text image super-resolution (TSR) aims to recover high-resolution text images from degraded low-resolution observations while preserving both visual details and character-level semantics~\citep{zhao2021pcan}. This challenge becomes more pronounced in the blind setting, where the degradation process is unknown. Unlike natural-image super-resolution, TSR must preserve not only perceptual quality but also character-level semantic accuracy, as a single missing, merged, or hallucinated stroke may lead to an incorrect transcription even when the reconstructed image appears visually sharp. Therefore, TSR models need to recover fine-grained visual structures while maintaining semantic fidelity.

\begin{figure}[t]
\centering
\includegraphics[width=1\columnwidth]{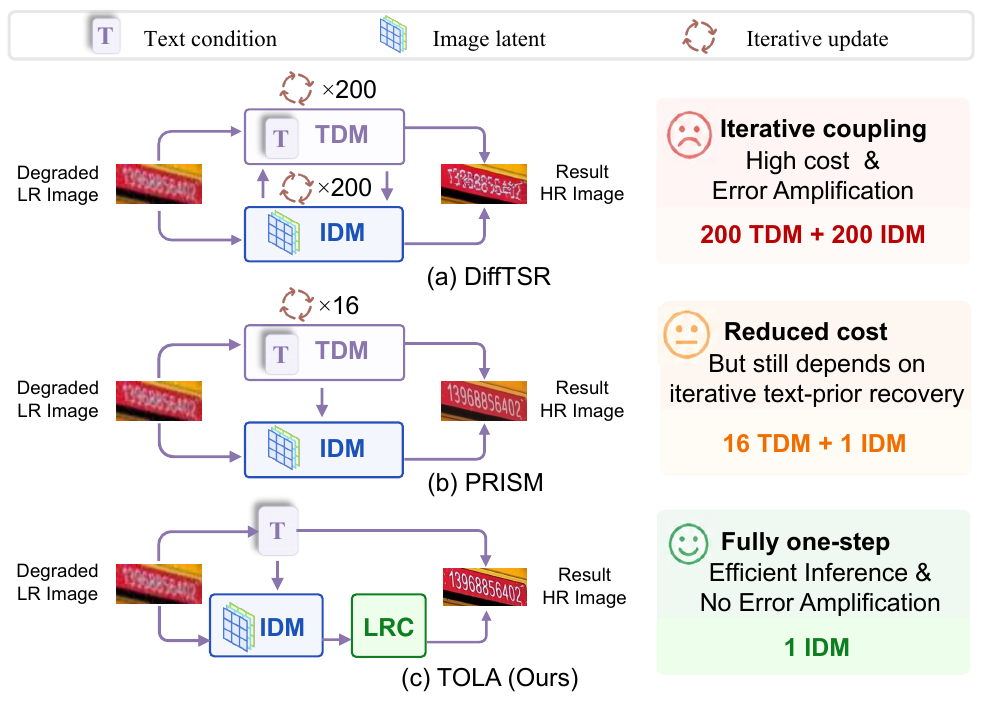}
\caption{Comparison of three different diffusion-based TSR inference paradigms. (a) DiffTSR performs 200-step iterative text-prior and image restoration. (b) PRISM uses one-step image reconstruction with 16-step text-prior recovery. (c) TOLA constructs the text condition once, evaluates IDM once, and applies a Latent Residual Correction (LRC) for clean-latent correction. Prior iterative text-image refinement may repeatedly propagate inaccurate semantic cues through multiple denoising stages, leading to accumulated recognition errors. In contrast, TOLA avoids error amplification by designing a one-step text-conditioned IDM.}
\label{fig:method_comparison}
\end{figure}

Early TSR methods enhance reconstruction by incorporating sequence modeling, text-aware attention, recognition guidance, and character-structure priors~\citep{wang2020tsrn,ma2022tatt,ma2023tpgsr,li2023marconet}. These methods perform well when the low-resolution input still contains reliable structural or semantic cues. Under severe degradations, however, missing and merged strokes make both visual reconstruction and text recognition highly ambiguous. Diffusion models have recently emerged as a promising solution because their generative priors can synthesize plausible high-frequency details that deterministic restoration networks often fail to recover~\citep{ho2020ddpm,rombach2022ldm,kawar2022ddrm,saharia2023sr3,wang2024stablesr}. Existing diffusion-based TSR methods typically rely on multi-step prediction of either the high-resolution image or its text-aware prior~\citep{zhang2024difftsr,xu2026prism}, as shown in \figref{fig:method_comparison}. Although iterative refinement improves generative capacity, it also introduces two fundamental limitations. First, repeatedly evaluating image and text networks incurs computational cost and inference latency. Second, the text prior used to guide reconstruction is itself inferred from the degraded observation and may therefore be incorrect~\citep{yang2024moreless}. When an erroneous prior is repeatedly injected into the denoising process, image and text predictions may reinforce each other, progressively amplifying an early recognition error into a sharp yet semantically incorrect character.

Directly reducing multi-step diffusion-based TSR to a single image-denoising step is non-trivial due to two challenges. First, reliable text guidance must be constructed in a single pass without repeatedly injecting uncertain OCR predictions into image reconstruction. Second, the clean latent estimated from one high-noise prediction inevitably contains structured errors, especially around thin strokes, character boundaries, and ambiguous local patterns, which can no longer be corrected by subsequent reverse updates. 
Introducing another iterative prior-recovery or refinement process would merely shift, rather than eliminate, the computational burden. The key is therefore to combine one-time semantic conditioning with explicit, non-iterative latent residual correction.

Based on this observation, we propose TOLA, a Text-aware One-step Latent Adaptation framework. Our central insight is that semantic conditioning and latent residual correction need not be coupled along the same reverse trajectory. 
Instead, reliable semantic evidence can be selected once before image prediction, while structured reconstruction errors can be modeled explicitly afterward in the clean-latent space. As illustrated in \figref{fig:method-overview}, TOLA consists of two complementary modules. First, the confidence-weighted text conditioning module uses frozen TransOCR to predict text tokens and token-wise confidence scores, then constructs the semantic condition once via the frozen MoM module. By suppressing uncertain OCR predictions, this module reduces the risk that an unreliable text hypothesis will dominate image reconstruction. Second, the lightweight latent residual correction module explicitly estimates the structured residual remaining in the initial clean latent and restores missing or distorted stroke details without introducing another reverse trajectory.

Our main contributions are summarized as follows:
\begin{itemize}
    \item We propose TOLA, which reformulates the original multi-step image-text diffusion pipeline into a non-iterative framework with one-shot text conditioning and one-step latent prediction.
    \item We introduce confidence-weighted text conditioning, which constructs the semantic condition once and suppresses unreliable character priors using OCR confidence scores. This prevents erroneous text information from being repeatedly injected and amplified during denoising, yielding reliability-aware semantic guidance. 
    \item Extensive experiments demonstrate that TOLA achieves state-of-the-art performance across all reported metrics on CTR-TSR-Test (\(\times4\)) and RealCE-200. TOLA outperforms the official 200-step DiffTSR in all nine metrics, improving PSNR by 4.2137 dB while reducing FID by 12.2271. Meanwhile, it shortens the average inference time from 9.9548s to 0.1319s, achieving a 75.5$\times$ speedup.
\end{itemize}

\section{Related Work}

\begingroup
\emergencystretch=1em

\noindent\textbf{Text-Aware Scene Text Image Super-Resolution.}
Generic SR optimizes visual fidelity without preserving character identity~\citep{dong2016srcnn,wang2018esrgan,chen2022nafnet,liang2021swinir}. TSR models strokes, sequences, layout, deformation, and location~\citep{wang2020tsrn,chen2021tbsrn,ma2022tatt,chen2022textgestalt,zhao2022c3stisr,zhu2023tsan,guo2023lemma,wei2025glyphsr,tomyenrique2024sgenet}. Recognition, style, and structure priors provide guidance~\citep{ma2023tpgsr,li2023marconet,zhu2023dpmn,park2025ncap,yuan2025stylesrn}, but cues extracted from degraded inputs remain unreliable~\citep{kong2024garden}.

\noindent\textbf{Diffusion-Based Text Image Restoration.}
Diffusion TSR conditions denoising on masks, LR text, recognition, or character embeddings~\citep{liu2025textdiff,zhou2024rgdiffsr,noguchi2024tcdm,singh2024dcdm}. DiffTSR couples IDM and TDM through MoM; Boosting DiffTSR adds mixed training, ResShift, cross-attention, and confidence weighting~\citep{zhang2024difftsr,pan2025boostingdifftsr}. TextSR, TEXTS-Diff, and TADiSR address multilingual or real-world restoration~\citep{ye2025textsr,he2026textsdiff,hu2025tadisr}; DualTSR and TeReDiff combine text diffusion or recognition with restoration~\citep{niu2026dualtsr,min2026terediff}. These methods retain iterative refinement.

\noindent\textbf{One-Step Diffusion Restoration.}
DDIM, ResShift, SinSR, and AddSR shorten or distill diffusion trajectories~\citep{song2021ddim,yue2023resshift,wang2024sinsr,tai2026addsr,dong2025tsdsr}; OSEDiff uses LoRA and score distillation~\citep{hu2022lora,wu2024osediff}, while FiDeSR combines frequency injection with latent residual refinement~\citep{kim2026fidesr}. These methods target generic SR rather than the coupled DiffTSR process. PRISM is text-specific but uses $K=16$ Euler steps for prior recovery~\citep{xu2026prism}. TOLA instead adapts pretrained DiffTSR to one IDM evaluation by constructing confidence-weighted TransOCR--MoM conditioning once and correcting the latent after noise-to-clean conversion.

\endgroup

\section{Method}

\begin{figure*}[t]
\centering
\includegraphics[width=1\textwidth]{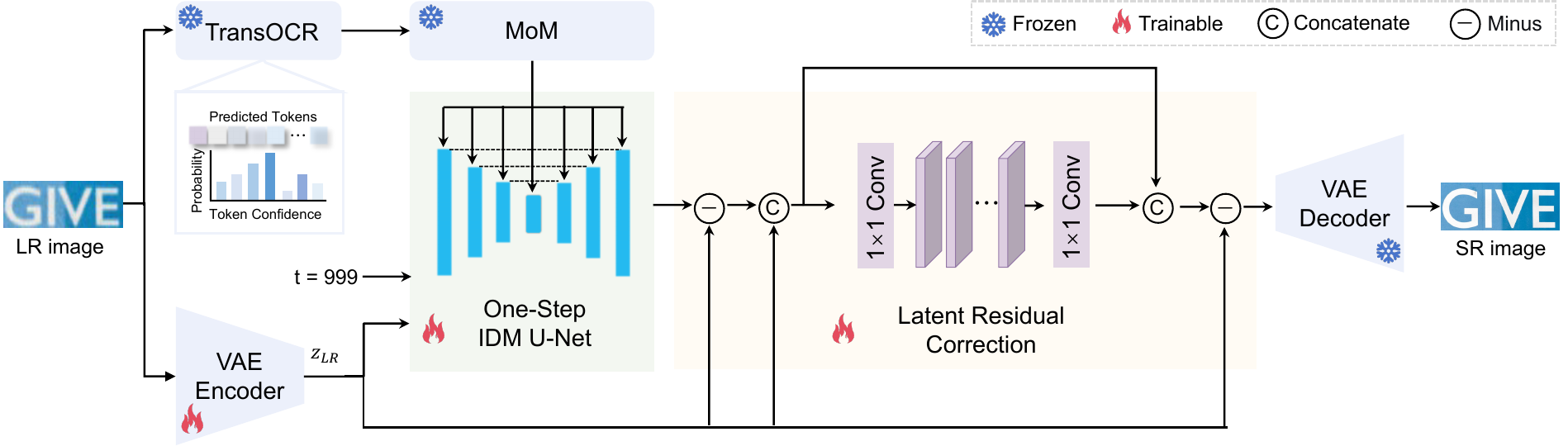}
\caption{TOLA architecture. The LR image is encoded and q-sampled at fixed $t=999$, while confidence-weighted TransOCR tokens and latent features form one MoM condition. A LoRA-adapted IDM predicts noise once, and LRC corrects the clean latent before frozen VAE decoding. Snowflakes and flames denote frozen and trainable components, respectively.}
\label{fig:method-overview}
\end{figure*}

\subsection{Overview}

As shown in \figref{fig:method-overview}, TOLA converts iterative DiffTSR into a single-pass restoration path. TransOCR predicts tokens and confidence scores, while the LoRA-adapted VAE encoder maps the LR image to $\vzLR$ and q-sampling obtains $\rvzt$ at $t=999$. MoM fuses confidence-weighted token embeddings with $[\vzLR,\rvzt]$ to construct $\tCCond$ once. Conditioned on $\tCCond$, the LoRA-adapted IDM performs one noise prediction, which is converted into the initial clean latent $\rvzZero$. LRC then corrects the residual $\vzLR-\rvzZero$ before VAE decoding. Only LoRA and LRC are optimized, and inference requires neither TDM nor iterative reverse updates.

\subsection{Text-Aware One-Step Adaptation}

\paragraph{Fixed-Timestep q-Sampling.}
The LoRA-adapted VAE encoder first encodes the degraded input image $\vxLR$ into an unscaled latent representation $\vzLR$,

\begin{equation}
\label{eq:vae-encoding}
\vzLR=\VAEEnc(\vxLR).
\end{equation}

Following DiffTSR q-sampling, we sample a noisy latent at the fixed, zero-indexed timestep $t=999$:

\begin{equation}
\label{eq:forward-noise}
\rvzt=\sqrt{\alphaBarT}\,\vzLR
+\sqrt{1-\alphaBarT}\,\rvepsilon,
\qquad \rvepsilon\sim\Normal(\vzero,\mI),
\end{equation}

where $\alphaBarT=\prod_{\tau=0}^{t}(1-\beta_\tau)$ follows the 1000-step linear schedule, with $\beta_0=0.0015$ and $\beta_{999}=0.0205$. We draw one Gaussian sample per input and supply $\vzLR$ and $\rvzt$ to IDM on the VAE latent scale. Unlike TOLA, OSEDiff feeds the unnoised LR latent to its one-step U-Net~\citep{wu2024osediff}.

\paragraph{Confidence-Weighted Text Conditioning.}
The frozen TransOCR instance used for conditioning predicts a token sequence and the corresponding token-wise confidence scores,

\begin{equation}
\label{eq:transocr-prediction}
\begin{aligned}
(\vcNative,\valpha)&=\TransOCR(\vxLR),\\
\vcPred&=\TextMap\!\left(\TextNorm(\vcNative)\right).
\end{aligned}
\end{equation}

$\TextNorm$ normalizes characters to Simplified Chinese; $\TextMap$ maps them to the IDM vocabulary with $\valpha$ alignment. Following Boosting DiffTSR~\citep{pan2025boostingdifftsr}, confidence-weighted embeddings enter MoM, which computes

\begin{equation}
\label{eq:mom-conditioning}
[\tICond,\tCCond]
=\gM\!\left(
s\,[\vzLR,\rvzt],
\valpha\odot\TokenEmb(\vcPred),
t
\right),
\end{equation}

where $s=0.18215$ scales image latents and $\odot$ broadcasts token confidences over embeddings. IDM uses only $\tCCond$ for cross-attention; $\tICond$ is discarded. Confidence is neither calibrated nor thresholded, and MoM runs once.

\paragraph{One-Step Clean-Latent Reconstruction.}
The LoRA-adapted IDM receives channel-wise concatenated noisy and LR latents and predicts

\begin{equation}
\label{eq:idm-prediction}
\rvepsilonTheta
=\fTheta\bigl([\rvzt,\vzLR],t,\tCCond\bigr),
\end{equation}

where $[\cdot,\cdot]$ is channel-wise concatenation and $\vtheta$ combines frozen base and trainable LoRA parameters. Since DiffTSR predicts noise, the clean latent is

\begin{equation}
\label{eq:clean-latent}
\rvzZero
=\frac{\rvzt-\sqrt{1-\alphaBarT}\,\rvepsilonTheta}
{\sqrt{\alphaBarT}}.
\end{equation}

No reverse update follows \eqref{eq:clean-latent}; $\rvzZero$ is passed directly to LRC. \Algref{alg:onestep-inference} summarizes inference.

\begin{algorithm}[t]
\caption{TOLA one-step inference.}
\label{alg:onestep-inference}
\begin{algorithmic}[1]
\renewcommand{\algorithmicensure}{\textbf{Output:}}
\REQUIRE LR image $\vxLR$; fixed $t=999$
\ENSURE Restored image $\rvxSR$
\STATE $\vzLR\leftarrow\VAEEnc(\vxLR)$; $(\vcNative,\valpha)\leftarrow\TransOCR(\vxLR)$
\STATE $\vcPred\leftarrow\TextMap(\TextNorm(\vcNative))$
\STATE Draw $\rvepsilon\sim\Normal(\vzero,\mI)$; construct $\rvzt$ with \eqref{eq:forward-noise}
\STATE $[\tICond,\tCCond]\leftarrow\gM(s\,[\vzLR,\rvzt],\valpha\odot\TokenEmb(\vcPred),t)$
\STATE $\rvepsilonTheta\leftarrow\fTheta([\rvzt,\vzLR],t,\tCCond)$
\STATE Reconstruct $\rvzZero$ with \eqref{eq:clean-latent}
\STATE $\rvr\leftarrow\vzLR-\rvzZero$; $\DeltaRvr\leftarrow\gPhi([\vzLR,\rvr])$
\STATE $\rvzZero^{\mathrm{corr}}\leftarrow \vzLR-(\rvr+\DeltaRvr)$; $\rvxSR\leftarrow \VAEDec(s^{-1}\rvzZero^{\mathrm{corr}})$
\end{algorithmic}
\end{algorithm}

TOLA evaluates the VAE encoder/decoder, TransOCR, MoM, IDM, and LRC once each while omitting TDM; DiffTSR evaluates IDM and TDM 200 times each.

\newcommand{\MainResultsTable}{%
\begin{table*}[t]
\centering
\begin{adjustbox}{max width=\linewidth}
\begin{tabular}{llcccccccc}
\toprule
Method & PSNR$\uparrow$ & SSIM$\uparrow$ & LPIPS$\downarrow$ & DISTS$\downarrow$ & FID$\downarrow$ & T-ACC$\uparrow$ & T-NED$\uparrow$ & P-ACC$\uparrow$ & P-NED$\uparrow$ \\
\midrule
SRCNN    & 22.0288 & 0.5860 & 0.6021 & 0.3886 & 156.9406 & 0.4255 & 0.6107 & 0.3303 & 0.5279 \\
ESRGAN   & 21.7741 & 0.5588 & 0.6455 & 0.3979 & 155.7055 & 0.4188 & 0.6061 & 0.3287 & 0.5273 \\
NAFNet   & 21.5568 & 0.5678 & 0.5924 & 0.3776 & 144.9606 & 0.4187 & 0.5998 & 0.2998 & 0.4870 \\
TSRN     & 20.4381 & 0.5840 & 0.6768 & 0.3926 & 140.2070 & 0.3307 & 0.5192 & 0.2238 & 0.4196 \\
TBSRN    & 20.6883 & 0.5840 & 0.6510 & 0.3918 & 137.1033 & 0.3539 & 0.5425 & 0.2574 & 0.4563 \\
TATT     & 20.9523 & 0.5846 & 0.6614 & 0.3896 & 125.6047 & 0.3558 & 0.5450 & 0.2491 & 0.4466 \\
MARCONet & 21.3460 & 0.6132 & 0.4386 & 0.3566 & 102.7384 & 0.3895 & 0.5735 & 0.2796 & 0.4600 \\
PRISM    & 22.08\mbox{-}\mbox{-} & -- & 0.2314 & -- & 12.57\mbox{-}\mbox{-} & 0.4212 & 0.6644 & -- & -- \\
DiffTSR  & 20.5958 & 0.6261 & 0.3103 & 0.2428 & 24.9120 & 0.4645 & 0.6519 & 0.3970 & 0.5873 \\
TeReDiff & 19.6333 & 0.5125 & 0.4714 & 0.3466 & 77.0387 & 0.2893 & 0.4491 & 0.1972 & 0.3592 \\
\midrule
Ours     & \textbf{24.8095} & \textbf{0.7235} & \textbf{0.2288} & \textbf{0.1953} & \textbf{12.6849} & \textbf{0.5401} & \textbf{0.7370} & \textbf{0.4626} & \textbf{0.6585} \\
\bottomrule
\end{tabular}
\end{adjustbox}
\caption{Results on 8,089 CTR-TSR-Test $\times4$ images. Non-diffusion methods are retrained on the CTR-TSR split, and all unified-protocol outputs share one evaluation implementation. T/P denote TransOCR/PaddleOCR; bold marks the best unified-protocol result. PRISM~\citep{xu2026prism} is a BTL-test $\times4$/PP-OCRv5 cross-protocol reference; dashes preserve its reported precision. \textit{Note: ``--'' denotes unavailable values. PRISM results are from its arXiv paper due to unavailable code.}}
\label{tab:main-comparison}
\end{table*}
}

\MainResultsTable

\newcommand{\CTRQualitativeFigure}{%
\begin{figure*}[t]
\centering
\includegraphics[
  width=\textwidth,
  height=0.442\textwidth
]{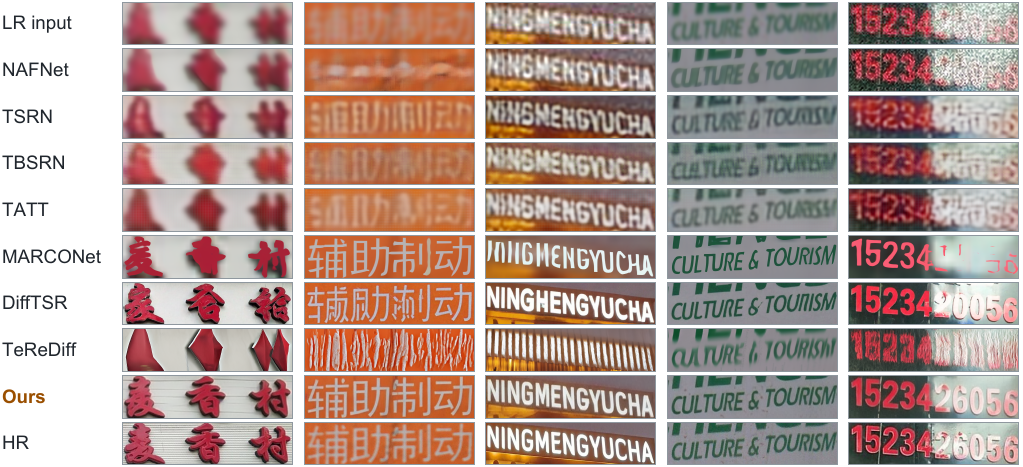}
\caption{Comparison on five selected CTR-TSR-Test $\times4$ images covering Chinese, English, and numeric text. Each column presents an aligned test sample, and the rows show the LR input, displayed baselines, Ours, and the HR reference.}
\label{fig:qualitative-ctr}
\end{figure*}
}

\subsection{Latent Residual Correction}

A FiDeSR-inspired LRC corrects local errors after noise-to-clean conversion~\citep{kim2026fidesr}. FiDeSR refines a directly predicted residual, whereas TOLA corrects the clean-latent residual without changing semantic conditioning.

Given the initial clean-latent estimate $\rvzZero$, we define
\begin{equation}
\label{eq:lrrb-residual}
\rvr=\vzLR-\rvzZero.
\end{equation}
From the concatenated input $[\vzLR,\rvr]$, LRC predicts
\begin{equation}
\label{eq:lrrb-correction}
\DeltaRvr=\gPhi([\vzLR,\rvr]),
\end{equation}
The trainable $\gPhi$ uses a $1\times1$ projection from 6 to 32 channels, one RRDB of three five-layer RDBs with $3\times3$ convolutions, and a $1\times1$ projection to 3 channels, with growth width 16 and residual scaling 0.2.

The corrected residual and latent are
\begin{equation}
\label{eq:lrrb-corrected-latent}
\rvr^{\mathrm{corr}}=\rvr+\DeltaRvr,
\qquad
\rvzZero^{\mathrm{corr}}=\vzLR-\rvr^{\mathrm{corr}}.
\end{equation}
The frozen VAE decoder then reconstructs
\begin{equation}
\label{eq:vae-decoding}
\rvxSR=\VAEDec\!\left(s^{-1}\rvzZero^{\mathrm{corr}}\right),
\end{equation}
The inherited scale is $s=0.18215$. Zero-initializing the LRC output projection initially preserves $\rvzZero$ and learns only the residual correction.

\subsection{Training Objective}

TOLA uses pixel, perceptual, and recognition objectives:
\begin{equation}
\label{eq:training-objective}
\begin{aligned}
\Ls ={}&
\Lpixel(\rvxSR,\vxHR)
+\Lpercep(\rvxSR,\vxHR) \\
&+0.02\,\Locr\!\left(\OCR(\rvxSR),\vyGT\right).
\end{aligned}
\end{equation}
$\vxHR$ and $\vyGT$ are the ground-truth image and transcription; $\Lpixel$ is $\ell_1$, $\Lpercep$ is LPIPS-VGG, and $\Locr$ is token-level cross-entropy. Two frozen TransOCR instances provide conditioning and OCR supervision; OCR gradients update LoRA and LRC. Conditioning uses the IDM vocabulary, whereas $\Locr$ retains the native TransOCR labels. No latent $\ell_1$, noise-prediction, distillation, or teacher loss is used.

\section{Experiments}

\subsection{Experimental Setup}

\paragraph{Datasets.} We use 62,371 training and 1,273 validation images from the CTR-TSR preprocessing set~\citep{zhang2024difftsr}. LR inputs are generated online with Real-ESRGAN or BSRGAN degradations at scales $\{1,2,4\}$~\citep{wang2021realesrgan,zhang2021bsrgan}; all images use a $128\times512$ canvas. Evaluation uses all 8,089 pairs in CTR-TSR-Test $\times4$.

For real-image generalization, we use two complementary test sets. RealCE-200 contains 200 filtered and deduplicated pairs from the RealCE validation split~\citep{ma2023realce}, using matched official $13\,\mathrm{mm}$ and $52\,\mathrm{mm}$ crops as LR and reference images. We construct RT50 to extend text-category and acquisition-condition coverage beyond existing paired benchmarks. Its 50 images, collected through multiple routes, include 17 Chinese, 17 English, and 16 numeric samples spanning varied blur, illumination, viewpoints, resolutions, and backgrounds. A predefined output-independent protocol screens content validity, geometry, degradation difficulty, quality, and duplication. Neither set is used for training, validation, or checkpoint selection.

\paragraph{Evaluation Protocol and Metrics.} 
To ensure a fair comparison, we adapt all non-diffusion methods to $\times4$ restoration and retrain them on the same CTR-TSR training split. All unified-protocol methods are evaluated on the same 8,089-pair manifest at $128\times512$ using identical filename matching and metric implementations. We report luminance-channel PSNR and SSIM without border cropping~\citep{wang2004ssim}, LPIPS-AlexNet~\citep{zhang2018lpips}, DISTS~\citep{ding2022dists}, and FID~\citep{heusel2017fid}; training instead uses LPIPS-VGG. TransOCR predictions undergo full-to-half-width and simplified-Chinese conversion and whitespace removal while preserving case. ACC denotes exact-match accuracy, and NED denotes mean normalized edit-distance similarity. Independent PaddleOCR ACC and NED~\citep{du2020ppocr} provide a recognizer cross-check.

\paragraph{Compared Methods.} 
We compare TOLA with SRCNN, ESRGAN, NAFNet, TSRN, TBSRN, TATT, MARCONet, DiffTSR, and TeReDiff, and additionally include the published PRISM results as a cross-protocol reference.

\paragraph{Implementation Details.}
Starting from the official CTR-trained DiffTSR checkpoint, we optimize rank-4 LoRA and LRC for 100K steps on four NVIDIA RTX PRO 6000 GPUs with batch size 16 per GPU. FP32 AdamW uses learning rate $5\times10^{-5}$, $(\beta_1,\beta_2)=(0.9,0.999)$, weight decay 0.01, 500-step linear warm-up, and gradient clipping at 1.0. Training takes about 55 hours, and validation selects the checkpoint. Loss weights are 1, 1, and 0.02. Each image is restored once without sample selection or averaging.

\CTRQualitativeFigure

\paragraph{Diagnostic Protocol.} 
All trainable variants in \Tabref{tab:component-ablation} share CTR data, DiffTSR initialization, data order, optimization, and the 8,089-image evaluation. Conditioning controls alter only the fixed checkpoint input; GT prior is an oracle. Paired comparisons use identical per-image noise.

\subsection{Main Results}

\paragraph{Quantitative Comparisons.}
\Tabref{tab:main-comparison} shows that TOLA surpasses DiffTSR in all nine metrics, gaining 4.2137~dB PSNR and 0.0756 TransOCR ACC while reducing FID by 12.2271. PaddleOCR confirms this trend: ACC/NED rise from 0.3970/0.5873 to 0.4626/0.6585.

\paragraph{CTR-TSR Qualitative Results.}
\Figref{fig:qualitative-ctr} shows more coherent strokes and closer character shapes and layouts across Chinese, English, and numeric examples.

\paragraph{Cross-Dataset Results.}
\Tabref{tab:cross-dataset} shows TOLA leading all six metrics, exceeding DiffTSR by 1.5953~dB PSNR, 4.50 points TransOCR ACC, and 8.00 points PaddleOCR ACC.

\begin{table}[t]
\centering
\small
\setlength{\tabcolsep}{1.0pt}
\begin{adjustbox}{max width=\linewidth}
\begin{tabular}{@{}lrrrrrr@{}}
\toprule
Method & PSNR$\uparrow$ & SSIM$\uparrow$ & T-ACC$\uparrow$ & T-NED$\uparrow$ & P-ACC$\uparrow$ & P-NED$\uparrow$ \\
\midrule
SRCNN & 19.1609 & 0.6024 & 0.3850 & 0.5837 & 0.4950 & 0.6670 \\
ESRGAN & 19.2062 & 0.5971 & 0.3850 & 0.5799 & 0.5000 & 0.6710 \\
NAFNet & 18.9038 & 0.5920 & 0.3650 & 0.5704 & 0.4750 & 0.6406 \\
TSRN & 18.6767 & 0.6009 & 0.2850 & 0.5103 & 0.3850 & 0.5649 \\
TBSRN & 18.9903 & 0.5947 & 0.3400 & 0.5364 & 0.3950 & 0.5980 \\
TATT & 18.9725 & 0.5954 & 0.3300 & 0.5291 & 0.4000 & 0.5970 \\
MARCONet & 18.3506 & 0.6198 & 0.3500 & 0.5408 & 0.3800 & 0.5972 \\
DiffTSR & 17.6287 & 0.5970 & 0.3500 & 0.5402 & 0.4300 & 0.6188 \\
TeReDiff & 17.8180 & 0.5272 & 0.2900 & 0.4826 & 0.3350 & 0.5508 \\
\midrule
Ours & \textbf{19.2240} & \textbf{0.6199} & \textbf{0.3950} & \textbf{0.5841} & \textbf{0.5100} & \textbf{0.6720} \\
\bottomrule
\end{tabular}
\end{adjustbox}
\caption{RealCE-200 cross-dataset results on the same 200 pairs. RealCE is excluded from training and checkpoint selection; T/P denote TransOCR/PaddleOCR.}
\label{tab:cross-dataset}
\end{table}

\begin{figure*}[t]
\centering
\includegraphics[
  width=\textwidth,
  height=0.473\textwidth
]{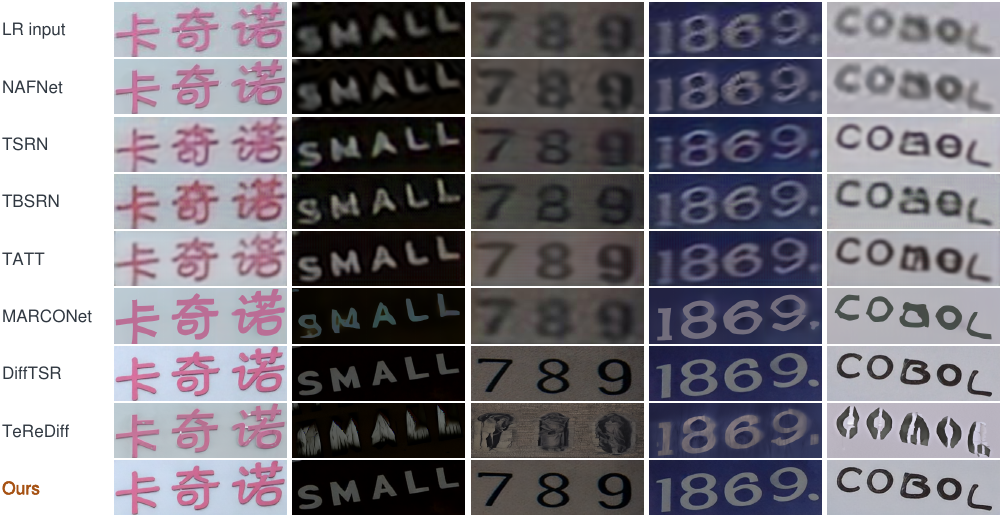}
\caption{RT50 comparison on Chinese, English, and numeric text under varied real degradations. Rows show LR inputs, displayed baselines, and Ours; examples cover blur, illumination, viewpoint, resolution, and background variations.}
\label{fig:qualitative-rt50}
\end{figure*}

\paragraph{Evaluation on RT50.}
\Figref{fig:qualitative-rt50} shows that TOLA better preserves strokes and character shapes on Chinese, English, and numeric text under varied real degradations.

\begin{figure*}[!t]
\centering
\includegraphics[width=0.96\textwidth]{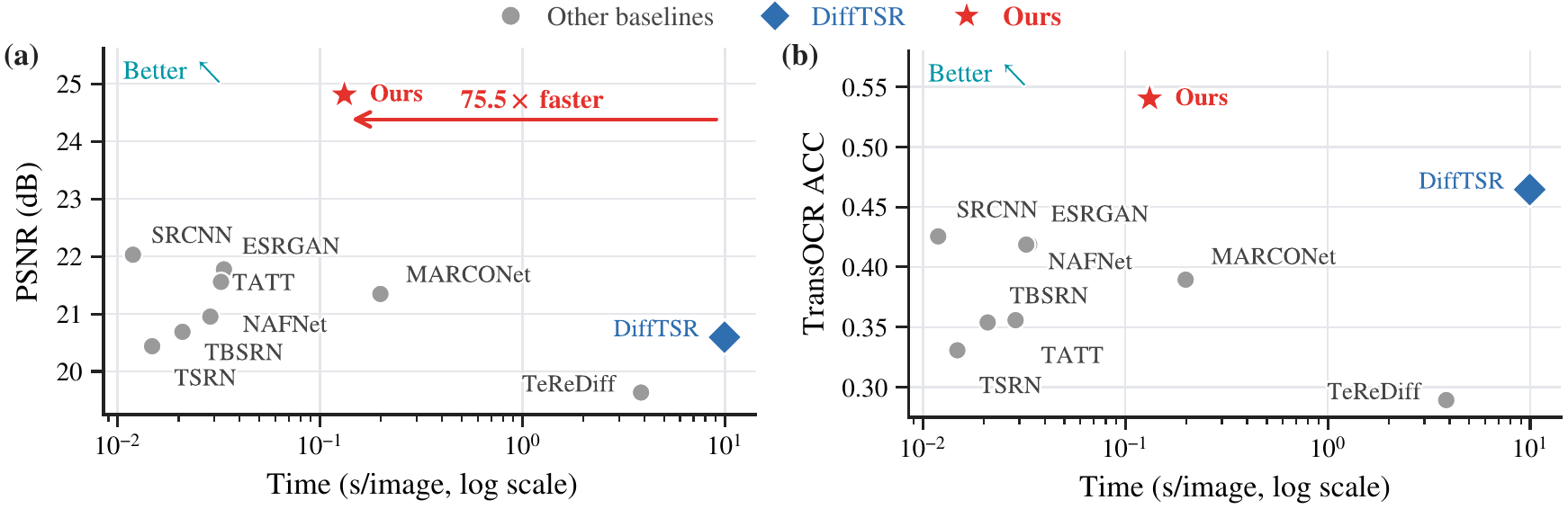}
\caption{Quality--latency comparison on CTR-TSR-Test $\times4$. Panels (a) and (b) plot PSNR and TransOCR ACC against synchronized batch-1 inference time measured on 100 fixed CTR inputs using one NVIDIA RTX PRO 6000.}

\label{fig:quality-latency}
\end{figure*}

\paragraph{Inference Cost and Latency.}
We use synchronized batch-1 timing on 100 fixed CTR images with one NVIDIA RTX PRO 6000, excluding loading. \Tabref{tab:efficiency} reports complete-trajectory MACs at $128\times512$ and restoration-module parameters excluding VAEs and auxiliary text modules. TOLA uses 297.497~G MACs and 0.1319 seconds per image, versus 59{,}213.970~G and 9.9548 seconds for DiffTSR, while its optimized and saved state is 2.46M parameters. \Figref{fig:quality-latency} visualizes the quality--latency trade-off.

\begin{table}[t]
\centering
\small
\setlength{\tabcolsep}{3.0pt}
\begin{adjustbox}{max width=\linewidth}
\begin{tabular}{@{}lrrr@{}}
\toprule
Method & \# Params. (M) & MACs (G) & Inference Time (s) \\
\midrule
SRCNN     &    0.057 &      3.748 & 0.0119 \\
ESRGAN    &   16.70  &     73.453 & 0.0335 \\
NAFNet    &   67.89  &     63.195 & 0.0324 \\
TSRN      &    2.68  &      0.904 & 0.0148 \\
TBSRN     &    3.21  &      2.535 & 0.0209 \\
TATT      &    7.61  &      1.269 & 0.0287 \\
MARCONet  &   87.90  &    466.178 & 0.1986 \\
DiffTSR   &  874.00  & 59{,}213.970 & 9.9548 \\
TeReDiff  & 1682.54  & 26{,}839.909 & 3.8499 \\
\midrule
Ours      &  876.23  &    297.497 & 0.1319 \\
\bottomrule
\end{tabular}
\end{adjustbox}
\caption{Restoration parameters, MACs, and batch-1 inference time on 100 CTR images. VAEs and text modules are excluded; TOLA stores 2.46M adapted parameters.}
\label{tab:efficiency}
\end{table}

\subsection{Ablations and Diagnostic Analyses}

\paragraph{Core Components.} 
\Tabref{tab:component-ablation} shows that LoRA drives one-step adaptation and LRC adds 0.3847~dB PSNR while reducing LPIPS/DISTS by 0.0121/0.0017; its ACC change is not significant ($p=0.584$). A parameter-matched plain correction loses 0.8249~dB PSNR and increases FID by 4.2491, so capacity alone does not explain the gain. IDM-only LoRA remains much closer to Full than VAE-only LoRA.

Removing MoM or bypassing its latent fusion degrades every metric. Relative to uniform confidence, confidence weighting adds 0.1908~dB PSNR and 0.0134/0.0138 TransOCR ACC/NED while reducing LPIPS/DISTS/FID by 0.0213/0.0160/4.2766. Null context confirms the value of predicted text, and the GT oracle adds another 0.0862 ACC and 0.0716 NED, revealing headroom from better priors.

\begin{figure*}[t]
\centering
\includegraphics[width=0.98\textwidth]{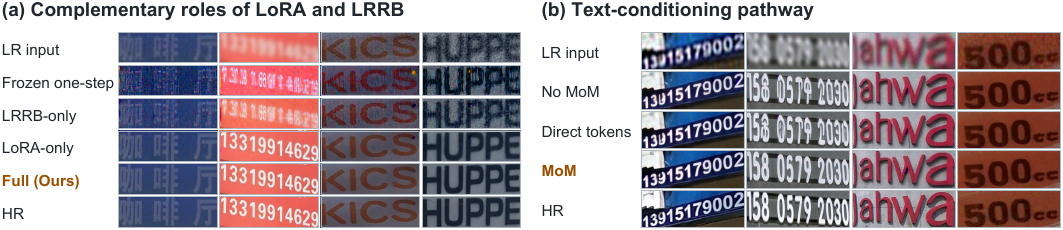}
\caption{Core component visualization. (a) LoRA enables one-step adaptation and LRC corrects local structure over frozen and single-module controls. (b) No MoM and direct-token conditioning cause character errors; LR and HR provide references.}
\label{fig:core-ablation-qualitative}
\end{figure*}

\begin{table*}[t]
\centering
\small
\renewcommand{\arraystretch}{1.03}
\begin{adjustbox}{max width=\linewidth}
\begin{tabular}{@{\extracolsep{\fill}}lrrrrrrrrr@{}}
\toprule
Configuration & PSNR$\uparrow$ & SSIM$\uparrow$ & LPIPS$\downarrow$ & DISTS$\downarrow$
& FID$\downarrow$ & T-ACC$\uparrow$ & T-NED$\uparrow$ & P-ACC$\uparrow$ & P-NED$\uparrow$ \\
\midrule
Frozen one-step                  & 20.0416 & 0.5342 & 0.5254 & 0.3464 & 119.4517 & 0.2276 & 0.3925 & 0.1278 & 0.2918 \\
LRC only                        & 21.9696 & 0.6021 & 0.4880 & 0.3301 & 110.0081 & 0.3785 & 0.5713 & 0.2440 & 0.4337 \\
LoRA only                        & 24.4220 & 0.7109 & 0.2410 & 0.1970 &  12.7420 & 0.5378 & 0.7337 & 0.4570 & 0.6564 \\
Full                             & \textbf{24.8095} & \textbf{0.7235} & \textbf{0.2288} & \textbf{0.1953} & \textbf{12.6849} & \textbf{0.5401} & \textbf{0.7370} & \textbf{0.4626} & \textbf{0.6585} \\
\midrule
Plain residual correction       & 23.9846 & 0.6991 & 0.2843 & 0.2180 &  16.9340 & 0.5080 & 0.7047 & 0.4256 & 0.6243 \\
No MoM                           & 23.9624 & 0.6984 & 0.2833 & 0.2170 &  16.1812 & 0.5060 & 0.7049 & 0.4117 & 0.6197 \\
Direct token condition          & 24.0026 & 0.6993 & 0.2846 & 0.2186 &  16.7650 & 0.5079 & 0.7038 & 0.4093 & 0.6185 \\
IDM LoRA only                   & 23.8812 & 0.6987 & 0.2844 & 0.2174 &  16.7593 & 0.5104 & 0.7058 & 0.4232 & 0.6254 \\
VAE LoRA only                   & 22.4861 & 0.6141 & 0.4635 & 0.3221 & 108.0600 & 0.3989 & 0.5926 & 0.2597 & 0.4505 \\
\midrule
Null context                    & 24.5495 & 0.7118 & 0.2695 & 0.2239 &  22.1417 & 0.5308 & 0.7280 & 0.4390 & 0.6438 \\
Uniform confidence              & 24.6159 & 0.7149 & 0.2502 & 0.2113 &  16.9615 & 0.5258 & 0.7224 & 0.4285 & 0.6356 \\
GT prior (oracle)               & 24.9148 & 0.7283 & 0.2216 & 0.1920 &  12.7565 & 0.6253 & 0.8079 & 0.5520 & 0.7387 \\
\bottomrule
\end{tabular}
\end{adjustbox}
\caption{Ablation on CTR-TSR-Test $\times4$. Full uses IDM/VAE LoRA, LRC, and confidence-aware MoM. Plain is parameter-matched; No MoM removes the text branch; direct tokens bypass fusion; GT is oracle; T/P denote TransOCR/PaddleOCR.}
\label{tab:component-ablation}
\end{table*}

\paragraph{Objective and Hyperparameter Diagnostics.} 
Rank 4 outperforms ranks 2 and 8. Small, default, and large LRCs use $(\text{hidden},\text{growth},\text{RRDBs})=(16,8,1)$, $(32,16,1)$, and $(32,16,2)$ with 0.045235M, 0.180323M, and 0.360323M parameters; the default gives the best trade-off. Removing $\ell_1$ most reduces PSNR, removing LPIPS most harms perceptual metrics, and OCR weight 0 lowers recognition; 0.05 brings no gain. Among tested timesteps, $t=999$ performs best.

\begin{figure}
\centering

\includegraphics[width=0.40\textwidth]{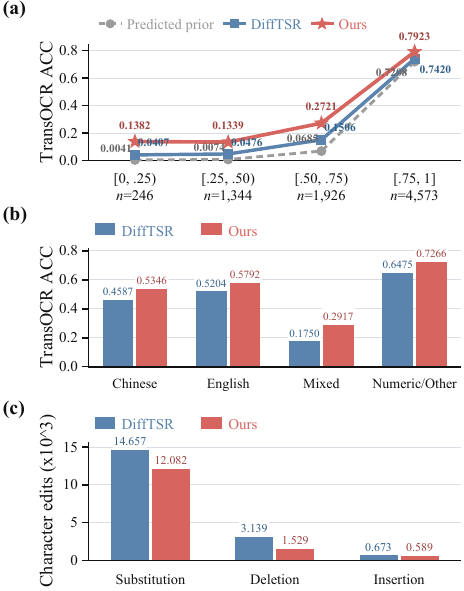}
\caption{Recognition diagnostics on CTR-TSR-Test $\times4$: (a) ACC across confidence bins, (b) ACC across text categories, and (c) character-level edit counts.}
\label{fig:recognition-diagnostics}
\end{figure}

\paragraph{Text-Prior Reliability and Error Analysis.} 
\Figref{fig:recognition-diagnostics} shows TOLA outperforms DiffTSR across confidence bins and text categories, reducing substitutions, deletions, and insertions. The exact-match gap between correct and incorrect priors identifies recognition errors as the main limitation.

\paragraph{Inference Controls and Stochastic Stability.} 
Controls favor $t=999$; zero noise and disabled VAE LoRA reduce quality, while LR-input and frozen-VAE controls rule out copying or VAE-only reconstruction. Across three draws, maximum standard deviations are 0.0019~dB PSNR and 0.0012 ACC.

\paragraph{Paired Uncertainty.} 
Paired bootstrap intervals exclude zero for LRC effects on PSNR, SSIM, LPIPS, and DISTS, but not NED; exact McNemar testing finds no significant ACC change. Predicted conditioning improves ACC over null context, and the GT-prior gap confirms prior errors.

\section{Conclusion}
We presented TOLA, a text-aware one-step adaptation of DiffTSR. TOLA constructs a confidence-aware TransOCR–MoM condition once, adapts IDM and the VAE encoder with rank-4 LoRA, and corrects clean-latent residual errors with LRRB without invoking TDM. It improves all nine CTR-TSR-Test metrics over DiffTSR while reducing IDM/TDM evaluations from 200/200 to 1/0 and latency from 9.9548 to 0.1319 seconds per image. The optimized and saved state contains only 2.46M parameters, while inference retains the pretrained backbone without introducing another iterative restoration trajectory. Component ablations and independent-OCR evaluation support the complementary roles of LoRA, LRRB, and confidence-aware conditioning. Results on RealCE-200 and RT50 further demonstrate generalization across paired camera observations, Chinese, English, and numeric text, and varied real capture conditions. These results establish efficient one-step diffusion TSR without iterative image–text refinement.

\bibliographystyle{plainnat}
\bibliography{main}

\clearpage
\appendix
\setcounter{secnumdepth}{2}
\setcounter{figure}{0}
\setcounter{table}{0}
\setcounter{equation}{0}
\renewcommand{\thefigure}{S\arabic{figure}}
\renewcommand{\thetable}{S\arabic{table}}
\renewcommand{\theequation}{S\arabic{equation}}
\renewcommand{\theHfigure}{supp.\arabic{figure}}
\renewcommand{\theHtable}{supp.\arabic{table}}
\renewcommand{\theHequation}{supp.\arabic{equation}}
\onecolumn
\section*{Appendices}
\input{supplementary}
\end{document}

%% file: math_commands.tex
\usepackage{amsmath,amsfonts,bm,upgreek}

\def\figref#1{figure~\ref{#1}}

\def\Figref#1{Figure~\ref{#1}}

\def\eqref#1{equation~\ref{#1}}

\def\Algref#1{Algorithm~\ref{#1}}

\def\rvepsilon{{\bm{\upepsilon}}}

\def\rvr{{\mathbf{r}}}

\def\rvx{{\mathbf{x}}}

\def\rvz{{\mathbf{z}}}

\def\vzero{{\bm{0}}}

\def\vtheta{{\bm{\theta}}}

\def\vphi{{\bm{\phi}}}

\def\valpha{{\bm{\alpha}}}

\def\vc{{\bm{c}}}

\def\vx{{\bm{x}}}

\def\vy{{\bm{y}}}

\def\vz{{\bm{z}}}

\def\mI{{\bm{I}}}

\DeclareMathAlphabet{\mathsfit}{\encodingdefault}{\sfdefault}{m}{sl}

\SetMathAlphabet{\mathsfit}{bold}{\encodingdefault}{\sfdefault}{bx}{n}

\def\gD{{\mathcal{D}}}

\def\gE{{\mathcal{E}}}

\def\gM{{\mathcal{M}}}

\newcommand{\Ls}{\mathcal{L}}

\def\Tabref#1{Table~\ref{#1}}

\newcommand{\vxLR}{\vx_{\mathrm{LR}}}
\newcommand{\vxHR}{\vx_{\mathrm{HR}}}
\newcommand{\vyGT}{\vy_{\mathrm{GT}}}
\newcommand{\vzLR}{\vz_{\mathrm{LR}}}
\newcommand{\vcPred}{\tilde{\vc}}
\newcommand{\vcNative}{\tilde{\vc}^{\mathrm{nat}}}

\newcommand{\rvxSR}{\rvx_{\mathrm{SR}}}
\newcommand{\rvzt}{\rvz_t}

\newcommand{\tICond}{\mathbf{I}_{\mathrm{cond}}}
\newcommand{\tCCond}{\mathbf{C}_{\mathrm{cond}}}

\newcommand{\rvepsilonTheta}{\hat{\rvepsilon}_{\vtheta}}

\newcommand{\DeltaRvr}{\Delta\rvr}
\newcommand{\rvzZero}{\hat{\rvz}_{0}}

\newcommand{\VAEEnc}{\gE_{\mathrm{VAE}}}
\newcommand{\VAEDec}{\gD_{\mathrm{VAE}}}

\newcommand{\alphaBarT}{\bar{\alpha}_t}
\newcommand{\fTheta}{f_{\vtheta}}
\newcommand{\gPhi}{g_{\vphi}}

\newcommand{\Lpixel}{\Ls_{1}}
\newcommand{\Lpercep}{\Ls_{\mathrm{LPIPS}}^{\mathrm{VGG}}}
\newcommand{\Locr}{\Ls_{\mathrm{OCR}}^{\mathrm{CE}}}

\newcommand{\Normal}{\mathcal{N}}
\DeclareMathOperator{\TransOCR}{TransOCR}
\DeclareMathOperator{\OCR}{OCR}
\DeclareMathOperator{\TokenEmb}{Emb}
\DeclareMathOperator{\TextNorm}{Norm}
\DeclareMathOperator{\TextMap}{Map}

%% file: authors.tex
\author{%
  \textbf{Yike Xu}$^{1,2}$\thanks{Work done during an internship at Shanghai Jiao Tong University.}\quad
  \textbf{Yue Shi}$^{1,3}$\quad
  \textbf{Yong Guo}$^{1}$\quad
  \textbf{Jiezhang Cao}$^{1}$\thanks{Corresponding author.}\\[0.4em]
  \normalsize $^{1}$Shanghai Jiao Tong University\quad
  $^{2}$University of Chinese Academy of Sciences\quad
  $^{3}$Shanghai AI Laboratory
}
\hypersetup{pdfauthor={Yike Xu; Yue Shi; Yong Guo; Jiezhang Cao}}

%% file: supplementary.tex
\setlength{\floatsep}{10pt}
\setlength{\intextsep}{10pt}
\captionsetup{font=small,skip=5pt}
\setlength{\parskip}{0pt}
\makeatletter
\renewcommand\normalsize{\@setfontsize\normalsize{9.5}{11}}
\renewcommand\small{\@setfontsize\small{8.5}{9.8}}
\renewcommand\paragraph{\@startsection{paragraph}{4}{\z@}%
  {1.2ex plus 0.3ex minus 0.2ex}{-0.7em}{\normalfont\normalsize\bfseries}}
\makeatother
\normalsize
\raggedbottom
\begin{multicols}{2}
This supplement provides implementation and evaluation details complementary
to the self-contained main paper. It clarifies the system boundary relative to
nearby diffusion methods, enumerates the modules executed by the one-step
route, specifies the saved adaptation state, datasets, metrics, and controlled
variants, and reports the exact values underlying the extended diagnostics.
It also presents additional ablations, inference controls, uncertainty
analysis, and qualitative comparisons.

\section{System and Comparison Boundaries}

\subsection{One-Step Inference Route}

For an input image, the VAE encoder first produces the unscaled three-channel
latent $z_{\mathrm{LR}}$. TOLA uses the zero-indexed timestep $t=999$ of the
1000-step linear schedule inherited from DiffTSR, whose endpoints are
$\beta_0=0.0015$ and $\beta_{999}=0.0205$. One Gaussian sample is drawn to
construct
\[
z_t=\sqrt{\bar{\alpha}_t}z_{\mathrm{LR}}
    +\sqrt{1-\bar{\alpha}_t}\epsilon .
\]
TransOCR is evaluated once. After Simplified-Chinese normalization, its token
embeddings are multiplied by the aligned raw token confidences and passed to
the frozen MoM together with $0.18215[z_{\mathrm{LR}},z_t]$. MoM computes both
inherited image and text conditions once, but the one-step IDM uses only the
text condition as cross-attention context.

IDM receives $[z_t,z_{\mathrm{LR}}]$ and predicts noise once. Because the
official DiffTSR checkpoint uses epsilon prediction, the initial clean latent
is reconstructed as
\[
\hat z_0 =
\frac{z_t-\sqrt{1-\bar{\alpha}_t}\hat\epsilon}
     {\sqrt{\bar{\alpha}_t}}.
\]
There is no reverse chain after this conversion. Latent residual correction
(LRC) uses a compact correction branch to correct the
LR-to-clean residual $r=z_{\mathrm{LR}}-\hat z_0$:
\[
\Delta r=g_\phi([z_{\mathrm{LR}},r]),\qquad
\hat z_{0,\mathrm{corr}}=z_{\mathrm{LR}}-(r+\Delta r).
\]
Finally, the frozen decoder maps
$\hat z_{0,\mathrm{corr}}/0.18215$ to $x_{\mathrm{SR}}$. Each image is
restored once, without multi-sample selection or averaging.

\section{Architecture and Saved State}

\subsection{LoRA Placement and Frozen Components}

Rank-4 LoRA adapters use $\alpha=4$ and zero dropout. They are inserted into
eligible linear and convolutional layers of IDM and the VAE encoder, including
its encoder-side $1\times1$ latent projection. The first input convolution of
IDM, the decoder-side latent projection, and the complete VAE decoder remain
unadapted.
TransOCR, MoM, all base IDM/VAE weights, and the loaded but unused TDM decoder
remain frozen. The conditioning TransOCR and the OCR-loss recognizer are
separate frozen instances initialized from the same checkpoint.

\setcounter{table}{2}

{
\setlength{\parindent}{0pt}
\begin{minipage}{\linewidth}
\centering
\small
\setlength{\tabcolsep}{4pt}
\begin{tabular*}{\linewidth}{@{\extracolsep{\fill}}lrr@{}}
\toprule
Trainable partition & Parameters & Saved? \\
\midrule
IDM LoRA & 2,027,520 & yes \\
VAE-encoder LoRA & 250,548 & yes \\
LRC & 180,323 & yes \\
\midrule
Total adaptation state & 2,458,391 & yes \\
\bottomrule
\end{tabular*}

\captionof{table}{Optimized and saved parameters.}
\label{tab:supp-adaptation-state}
\end{minipage}
\par}

\subsection{Latent Residual Correction}

TOLA implements LRC with a compact correction branch inspired by FiDeSR. It receives the
six-channel concatenation $[z_{\mathrm{LR}},r]$. A $1{\times}1$ input
projection maps 6 channels to 32, followed by one residual-in-residual dense
block (RRDB) and a $1{\times}1$ output projection from 32 channels to 3. The
RRDB contains three residual dense blocks; each residual dense block contains
five densely connected $3{\times}3$ convolutions with growth width 16.
Residual scaling is 0.2 at both dense-block and RRDB levels. The output
projection is zero initialized, so LRC initially leaves $\hat z_0$ unchanged.

Small, Default, and Large LRC use
$(h,g,n_{\mathrm{RRDB}})=(16,8,1)$, $(32,16,1)$, and $(32,16,2)$,
respectively, with 0.045235M, 0.180323M, and 0.360323M parameters. The
parameter-matched baseline uses a plain correction branch with a
$6{\rightarrow}30$ projection, 11 two-convolution residual blocks with
LeakyReLU and residual scaling 0.2, a $30{\rightarrow}30$ mixing layer, and a
zero-initialized $30{\rightarrow}3$ projection (0.180093M parameters).
\section{Training Protocol}

\paragraph{Computing Environment.}
Experiments run on Ubuntu 22.04 with an AMD EPYC 9J14 96-Core Processor,
503\,GiB of system memory, and NVIDIA RTX PRO 6000 Blackwell Server Edition
GPUs with 97,887\,MiB per GPU. The NVIDIA driver version is 580.142. The
primary environment uses Python 3.10.12, PyTorch 2.11.0+cu128,
Torchvision 0.26.0+cu128, CUDA 12.8, cuDNN 9.19, PyIQA 0.1.15.post2, and
LPIPS 0.1.4. The independent OCR evaluation uses PaddlePaddle 2.6.2 and
PaddleOCR 2.7.3. These package versions define the software environment used
for the reported results.

Training reads 63,644 high-quality CTR preprocessing images and applies a
deterministic local 98\%/2\% partition: 62,371 images for training and 1,273
for validation. This is a local partition rather than an official CTR split.
Each image is degraded online with equal probability by the Real-ESRGAN or
BSRGAN pipeline~\citep{wang2021realesrgan,zhang2021bsrgan}; the scale is drawn
from $\{1,2,4\}$. Both degraded inputs and targets use a $128{\times}512$
canvas.

The complete objective is
\[
\mathcal L =
\mathcal L_1(x_{\mathrm{SR}},x_{\mathrm{HR}})
+\mathcal L_{\mathrm{LPIPS\mbox{-}VGG}}(x_{\mathrm{SR}},x_{\mathrm{HR}})
+0.02\,\mathcal L_{\mathrm{OCR}}.
\]
The frozen OCR recognizer remains differentiable with respect to its image
input, so OCR-loss gradients reach LoRA and LRC. No latent $\ell_1$,
noise-prediction, score-distillation, VSD/CSD, or teacher-regularization loss
is used.

\end{multicols}

\clearpage
{\setlength{\parindent}{0pt}
\setcounter{table}{0}

{
\setlength{\parindent}{0pt}
\centering
\small
\setlength{\tabcolsep}{4.2pt}
\renewcommand{\arraystretch}{1.08}
\begin{tabular*}{\linewidth}{@{\extracolsep{\fill}}llll@{}}
\toprule
Method & Image route & Text route & Distinction \\
\midrule
DiffTSR
& Iterative IDM
& TDM--MoM iterations
& Coupled image/text diffusion \\
Boosting DiffTSR
& Iterative diffusion
& Confidence-guided TDM--MoM
& SR prior + progressive sampling \\
OSEDiff
& One-step latent
& Degradation prompt
& LoRA + VSD \\
FiDeSR
& One-step residual
& No OCR
& LRC + detail/frequency modules \\
PRISM
& Single-pass flow
& Rectified prior
& Flow matching + structure modeling \\
TADiSR
& Text-aware diffusion
& Text segmentation
& Joint segmentation decoders \\
TEXTS-Diff
& One-step latent
& Abstract + region cues
& Real-Texts + text-aware restoration \\
DualTSR
& Conditional flow
& Discrete diffusion
& Shared multimodal transformer \\
TOLA \textbf{(Ours)}
& One IDM call
& TransOCR--MoM once
& One-Step; no TDM \\
\bottomrule
\end{tabular*}

\captionof{table}{Mechanistic positioning of the closest diffusion routes.
Methods follow~\citep{zhang2024difftsr,pan2025boostingdifftsr,wu2024osediff,
kim2026fidesr,xu2026prism,hu2025tadisr,he2026textsdiff,niu2026dualtsr}.
The table distinguishes system boundaries.}
\label{tab:supp-closest-methods}
\par}
\vspace{8pt}

{
\setlength{\parindent}{0pt}
\centering
\small
\setlength{\tabcolsep}{5.0pt}
\begin{tabular*}{\linewidth}{@{\extracolsep{\fill}}lllr@{}}
\toprule
Component & Function & Trainable update & Calls/image \\
\midrule
TransOCR & token and confidence prediction & frozen & 1 \\
VAE encoder & LR latent encoding & LoRA & 1 \\
MoM & text-aware condition construction & frozen & 1 \\
IDM U-Net & one-step diffusion prediction & LoRA & 1 \\
LRC & latent residual correction & full branch & 1 \\
VAE decoder & corrected-latent decoding & frozen & 1 \\
\bottomrule
\end{tabular*}

\captionof{table}{Inference boundary of TOLA. TransOCR, MoM, and IDM are each
executed once; TDM is not executed.}
\label{tab:supp-inference-boundary}
\par}
\par}
\vspace{8pt}
\begin{multicols}{2}
\setcounter{table}{3}

{
\setlength{\parindent}{0pt}
\noindent\begin{minipage}{\linewidth}
\centering
\small
\renewcommand{\arraystretch}{1.08}
\setlength{\tabcolsep}{5pt}
\begin{tabular*}{\linewidth}{@{\extracolsep{\fill}}ll@{}}
\toprule
Item & Setting \\
\midrule
Initialization & official DiffTSR CTR checkpoint \\
Training steps & 100,000 \\
GPUs & 4$\times$ RTX PRO 6000 \\
Batch/GPU & 16 \\
Effective batch & 64 \\
Precision & FP32 \\
Training time & approximately 55 hours \\
Optimizer & AdamW \\
Learning rate & $5\times10^{-5}$ \\
Adam betas & $(0.9,0.999)$ \\
Weight decay & 0.01 \\
Warm-up & 500 linear steps \\
Gradient clipping & 1.0 \\
Selection & validation performance \\
\bottomrule
\end{tabular*}

\captionof{table}{Primary training configuration.}
\label{tab:supp-training}
\end{minipage}
\par}

\subsection{Hyperparameter Search and Selection}

Table~\ref{tab:supp-hyperparameter-search} summarizes the search. The final
architecture, objective, OCR-loss weight, and training timestep are selected
using validation performance. Most candidate checkpoints use validation LPIPS;
the no-LPIPS objective uses validation PSNR. The remaining optimization
settings are reported in Table~\ref{tab:supp-training}.

\section{Dataset Appendix and Evaluation Metrics}

\subsection{Dataset Inventory}

Table~\ref{tab:supp-dataset-inventory} summarizes the evaluation sets. The
primary benchmark is the complete 8,089-image CTR-TSR-Test $\times4$ set. All
methods under the unified protocol are filename-matched to the same references
and evaluated on a $128{\times}512$ output canvas.

\subsection{Motivation and Coverage}

Existing benchmarks do not individually cover all conditions required for
evaluating real scene-text restoration. CTR-TSR-Test provides a large paired
benchmark but primarily reflects synthetic degradations, whereas available
real-image datasets differ in language coverage, capture conditions,
annotation format, and reference availability. RealCE-200 provides a
paired protocol constructed from real multi-focal observations, while RT50 balances
Chinese, English, and numeric text collected through heterogeneous
acquisition routes. Together, these sets broaden the evaluated conditions.
\subsection{RealCE-200 Construction}

We construct RealCE-200 from the multi-focal images and official text
annotations in the RealCE validation split before evaluation. For
each official text annotation, identical bounding-box coordinates are applied
to the corresponding $13\,\mathrm{mm}$ and $52\,\mathrm{mm}$ focal images,
which serve as LR and reference observations. Both crops are bicubically
resized to $128{\times}512$ without geometric registration. Automatic
filtering considers transcription length and composition, bounding-box size,
aspect ratio, sharpness, and contrast. Duplicate candidates are removed using
transcription identity and perceptual dHash. Official transcriptions are
retained after whitespace cleaning, and no manual relabeling is performed.
The resulting records use fixed identifiers R001--R200. RealCE is not used for
training, validation, or checkpoint selection.

\end{multicols}

\clearpage
{\setlength{\parindent}{0pt}

{
\setlength{\parindent}{0pt}
\centering
\small
\setlength{\tabcolsep}{6pt}
\renewcommand{\arraystretch}{1.08}
\begin{tabular*}{\linewidth}{@{\extracolsep{\fill}}llcl@{}}
\toprule
Hyperparameter & Values examined & Count & Final setting \\
\midrule
LoRA $(r,\alpha)$ & $(2,2)$, $(4,4)$, $(8,8)$ & 3 & $(4,4)$ \\
LRC $(h,g,n_{\mathrm{RRDB}})$ & $(16,8,1)$, $(32,16,1)$, $(32,16,2)$ & 3 & $(32,16,1)$ \\
Reconstruction objective & Full, without $\ell_1$, without LPIPS & 3 & Full \\
OCR-loss weight & $\{0,0.02,0.05\}$ & 3 & 0.02 \\
Training timestep & $\{899,949,999\}$ & 3 & 999 \\
\bottomrule
\end{tabular*}

\captionof{table}{Hyperparameter and objective settings examined during
development and the final configuration used by TOLA.}
\label{tab:supp-hyperparameter-search}
\par}
\vspace{8pt}

{
\setlength{\parindent}{0pt}
\centering
\small
\setlength{\tabcolsep}{4pt}
\renewcommand{\arraystretch}{1.10}
\begin{tabular*}{\linewidth}{@{\extracolsep{\fill}}llll@{}}
\toprule
Set & Source & Size & Content and reference \\
\midrule
CTR-TSR-Test $\times4$
& Official CTR-TSR-Test
& 8,089 pairs
& Filename-aligned LR/HR images and transcriptions \\
RealCE-200
& RealCE multi-focal images~\citep{ma2023realce}
& 200 pairs
& Paired $13\,\mathrm{mm}$ LR/$52\,\mathrm{mm}$ reference crops \\
RT50
& Multi-source real images
& 50 inputs
& 17 Chinese, 17 English, and 16 numeric samples \\
\bottomrule
\end{tabular*}

\captionof{table}{Evaluation datasets. RealCE-200 and RT50 are constructed in this work.}
\label{tab:supp-dataset-inventory}
\par}
\par}
\vspace{8pt}
\begin{multicols}{2}
\subsection{RT50 Construction}

RT50 contains 50 real text inputs collected through multiple acquisition
routes. The fixed composition contains 17 Chinese samples, 17 English
samples, and 16 numeric samples; the difficulty partition contains 10 mild,
20 medium, and 20 hard samples. Candidate labels are restricted to
2--8 Chinese characters, 3--12 English letters, or 2--6 digits. Images must be at
least $24{\times}8$ pixels with aspect ratios between 2 and 10. After
canonical resizing, admissible candidates have gray mean in $[20,235]$, gray
standard deviation in $[10,80]$, Laplacian variance in $[2,1200]$, edge
density in $[0.01,0.65]$, and entropy of at least 3.0. Repeated
transcriptions are removed, and near duplicates are rejected at a dHash
Hamming-distance threshold of 7. The selection protocol is finalized before
model inference without consulting restoration or OCR outputs. Records use
fixed identifiers RT001--RT050 and are normalized to the same
$128{\times}512$ model-input canvas.

\subsection{Included Dataset Records}

The public data release will include the exact image files evaluated in this
work rather than requiring users to reconstruct them. For RealCE-200, it will
provide all 200 LR/reference pairs together with the pair manifest, transcriptions,
source filenames, and SHA-256 values. For RT50, it will provide all 50 input
images together with the complete manifest, labels, provenance table,
selection protocol, per-image geometry and quality statistics, category and
difficulty assignment, dHash, and SHA-256 values. The accompanying records
define the evaluated samples independently of model outputs and verify every
released file. Upon publication, these exact datasets and their metadata will
be publicly available under research-use terms documented with the release.

\subsection{Metric Implementations}

Predictions and references are matched by exact filename. PyIQA computes PSNR
and SSIM on luminance without border crop. Evaluation LPIPS uses the AlexNet
backbone, while training LPIPS uses VGG. DISTS and FID use PyIQA; FID is
computed once over the complete output and reference directories.

The primary recognizer is TransOCR. Predictions undergo full-width to
half-width conversion, Simplified-Chinese conversion, and whitespace removal;
case is preserved. Exact-match accuracy is denoted ACC. Per-image normalized
edit-distance similarity is
\[
\mathrm{NED}=1-
\frac{\mathrm{ED}(p,g)}
{\max(|p|,|g|,1)},
\]
and is averaged across the set. The independent cross-check uses PaddleOCR
2.7.3 with PaddlePaddle 2.6.2 and the Chinese PP-OCRv4 recognition model;
detection and orientation classification are disabled. This recognizer is not
used for conditioning, training, validation, or checkpoint selection.

\section{Extended Qualitative Comparisons}

\subsection{Additional RealCE Comparison}

Figure~\ref{fig:supp-qualitative-realce} provides additional paired RealCE-200
examples under the unified evaluation protocol.

\subsection{BTL Cross-Protocol Comparison}

\Figref{fig:supp-qualitative-btl} compares TOLA, DiffTSR, TeReDiff, and PRISM
on five BTL-test $\times4$ examples~\citep{xu2026prism}. The examples cover
Chinese text, English words, and numeric sequences.
\section{Controlled Variant Definitions}

All filename-paired comparisons use identical per-image Gaussian noise for the
two compared outputs. Effect-size intervals use 5,000 filename-paired
bootstrap resamples. Exact-match ACC is additionally tested with the exact
paired McNemar test. Table~\ref{tab:supp-variant-definitions} defines all
architecture and conditioning controls.

\end{multicols}
\clearpage

{
\setlength{\parindent}{0pt}
\centering
\includegraphics[width=0.94\textwidth]{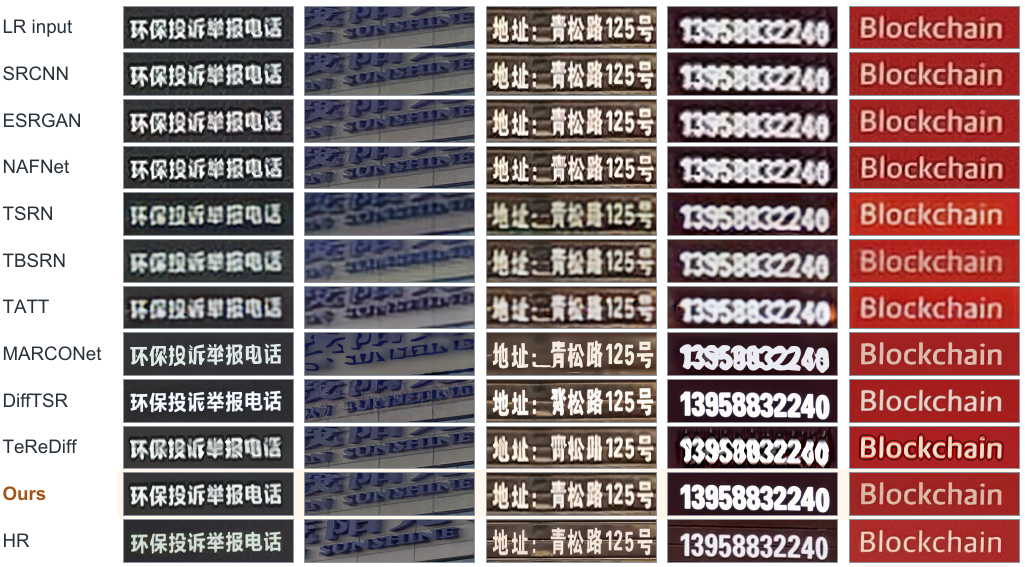}
\captionof{figure}{Qualitative comparison on five paired RealCE-200 examples.
Rows show the LR input, SRCNN, ESRGAN, NAFNet, TSRN, TBSRN, TATT, MARCONet,
DiffTSR, TeReDiff, Ours, and the HR reference.}
\label{fig:supp-qualitative-realce}
\par}
\vspace{10pt}

{
\setlength{\parindent}{0pt}
\centering
\includegraphics[width=0.94\textwidth]{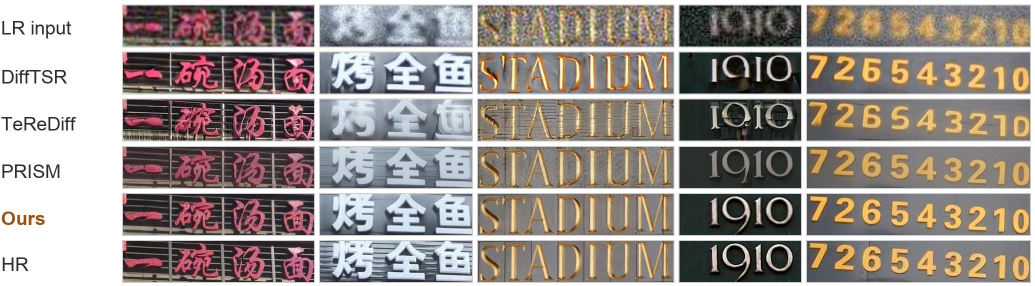}
\captionof{figure}{Cross-protocol comparison on five BTL-test $\times4$
examples. Rows show the LR input, DiffTSR,
\mbox{TeReDiff}~\citep{min2026terediff}, PRISM~\citep{xu2026prism}, Ours, and
GT. BTL-test combines real and rendered HQ sources with BSRGAN/Real-ESRGAN
degradations; the PRISM row is reproduced from the original paper.}
\label{fig:supp-qualitative-btl}
\par}
\vspace{10pt}

{
\setlength{\parindent}{0pt}
\begin{minipage}{\textwidth}
\centering
\small
\setlength{\tabcolsep}{8pt}
\renewcommand{\arraystretch}{1.08}
\begin{tabular*}{\linewidth}{@{\extracolsep{\fill}}p{0.18\linewidth}p{0.77\linewidth}@{}}
\toprule
Variant & Definition \\
\midrule
Frozen one-step
& Frozen DiffTSR pathway with one IDM call and no LoRA or LRC. \\
LRC only
& Default LRC is optimized while the pretrained image pathway remains frozen. \\
LoRA only
& Rank-4 IDM and VAE-encoder LoRA without LRC. \\
Full
& Rank-4 LoRA plus the default LRC and confidence-aware MoM condition. \\
No MoM
& The TransOCR--MoM conditioning branch is removed during training and inference. \\
Direct tokens
& Predicted token embeddings are supplied without MoM latent fusion. \\
Null context
& Fixed-checkpoint control retaining MoM with a null text context. \\
Uniform confidence
& Fixed-checkpoint control retaining predicted tokens and MoM with unit confidences. \\
GT prior
& Fixed-checkpoint oracle replacing predicted tokens with ground-truth text. \\
IDM LoRA only
& VAE-encoder LoRA is omitted; IDM LoRA and default LRC remain. \\
VAE LoRA only
& IDM LoRA is omitted; VAE-encoder LoRA and default LRC remain. \\
VAE LoRA off
& Inference disables all VAE-encoder LoRA, including the latent projection, while retaining IDM LoRA and LRC. \\
\bottomrule
\end{tabular*}

\captionof{table}{Architecture and conditioning controls used in the diagnostics.}
\label{tab:supp-variant-definitions}
\end{minipage}
\par}

\clearpage
{\setlength{\parindent}{0pt}

{
\setlength{\parindent}{0pt}
\begin{minipage}[t]{0.49\textwidth}
\centering
\small
\setlength{\tabcolsep}{1.25pt}
\renewcommand{\arraystretch}{1.08}
\begin{tabular*}{\linewidth}{@{\extracolsep{\fill}}lrrrrr@{}}
\toprule
Configuration & PSNR & SSIM & LPIPS & DISTS & FID \\
\midrule
Full & 24.8095 & .7235 & .2288 & .1953 & 12.6849 \\
Rank 2 & 23.9287 & .6979 & .2835 & .2168 & 16.8514 \\
Rank 8 & 23.9573 & .6993 & .2815 & .2179 & 17.2784 \\
LRC Small & 23.9506 & .6984 & .2870 & .2192 & 17.4977 \\
LRC Large & 23.9967 & .6990 & .2781 & .2154 & 15.6799 \\
No $\ell_1$ & 23.1142 & .6906 & .2801 & .2153 & 15.6182 \\
No LPIPS & 24.0026 & .6888 & .3820 & .2744 & 50.5338 \\
OCR wt. 0 & 23.9157 & .6997 & .2741 & .2132 & 16.0079 \\
OCR wt. .05 & 23.9290 & .6961 & .2955 & .2232 & 18.1809 \\
Train $t=899$ & 24.2532 & .7088 & .2571 & .2101 & 13.5299 \\
Train $t=949$ & 24.1818 & .7068 & .2644 & .2121 & 13.8379 \\
\bottomrule
\end{tabular*}

\captionof{table}{Image-quality diagnostics on CTR-TSR-Test $\times4$. The
reference uses rank 4, the default 0.180M LRC, OCR weight 0.02, and $t=999$.}
\label{tab:supp-sensitivity}
\end{minipage}
\hfill
\begin{minipage}[t]{0.49\textwidth}
\centering
\small
\setlength{\tabcolsep}{2.5pt}
\renewcommand{\arraystretch}{1.08}
\begin{tabular*}{\linewidth}{@{\extracolsep{\fill}}lrrrr@{}}
\toprule
Configuration & T-ACC & T-NED & P-ACC & P-NED \\
\midrule
Full & .5401 & .7370 & .4626 & .6585 \\
Rank 2 & .5138 & .7056 & .4253 & .6247 \\
Rank 8 & .5119 & .7052 & .4211 & .6224 \\
LRC Small & .5079 & .7038 & .4235 & .6235 \\
LRC Large & .5097 & .7054 & .4247 & .6240 \\
No $\ell_1$ & .5092 & .7031 & .4248 & .6237 \\
No LPIPS & .5044 & .6987 & .4015 & .6049 \\
OCR wt. 0 & .5025 & .6981 & .4253 & .6216 \\
OCR wt. .05 & .5096 & .7042 & .4217 & .6234 \\
Train $t=899$ & .5127 & .7063 & .4249 & .6266 \\
Train $t=949$ & .5107 & .7060 & .4271 & .6256 \\
\bottomrule
\end{tabular*}

\captionof{table}{Recognition diagnostics for the same configurations. Most
checkpoints use validation LPIPS; the no-LPIPS variant uses validation PSNR.}
\label{tab:supp-sensitivity-recognition}
\end{minipage}
\par}
\par}
\vspace{8pt}
\begin{multicols}{2}
\paragraph{Inference Runs.}
The main comparisons, ablations, and controls use one restored output per
image without multi-sample selection or output averaging. The
stochastic-stability analysis contains three inference runs and reports their
mean and standard deviation. Bootstrap resampling does not generate additional
restored outputs.

\section{Extended Ablation and Control Results}

\subsection{Objective and Hyperparameter Diagnostics}

Tables~\ref{tab:supp-sensitivity} and
\ref{tab:supp-sensitivity-recognition} show that rank 4 provides a better
overall balance than ranks 2 and 8. The default LRC also outperforms its smaller
and larger variants overall, indicating that additional capacity is not the
source of the gain. Removing $\ell_1$ causes the largest PSNR reduction,
whereas removing LPIPS most strongly degrades perceptual and distributional
metrics. An OCR weight of 0.02 outperforms both zero weight and 0.05, and
$t=999$ gives the strongest overall training result among the tested
timesteps.

\subsection{Prior-Stratified Performance}

Table~\ref{tab:supp-prior-performance} separates samples according to whether
the predicted TransOCR prior matches the ground truth.

\par

{
\setlength{\parindent}{0pt}
\noindent\begin{minipage}{\linewidth}
\centering
\small
\renewcommand{\arraystretch}{0.92}
\captionsetup{skip=2pt}
\setlength{\tabcolsep}{5pt}
\begin{tabular*}{\linewidth}{@{\extracolsep{\fill}}lrr@{}}
\toprule
Statistic & Correct prior & Wrong prior \\
\midrule
$N$             & 3,439  & 4,650 \\
Mean confidence & 0.9536 & 0.6015 \\
DiffTSR ACC     & 0.9514 & 0.1043 \\
TOLA ACC        & 0.9683 & 0.2217 \\
DiffTSR NED     & 0.9872 & 0.4038 \\
TOLA NED        & 0.9906 & 0.5482 \\
TOLA PSNR       & 26.2307 & 23.7535 \\
TOLA LPIPS      & 0.1652 & 0.2760 \\
\bottomrule
\end{tabular*}

\captionof{table}{Performance stratified by TransOCR prior correctness, with image and recognition metrics.}
\label{tab:supp-prior-performance}
\end{minipage}
\par}

\subsection{Inference Controls and Stochastic Stability}

Table~\ref{tab:supp-inference-stability} shows that $t=949$ slightly increases
PSNR and SSIM but worsens LPIPS, DISTS, and FID, while $t=899$ further
degrades perceptual and distributional quality. The default $t=999$ therefore
provides the best overall balance. Zero noise and disabled VAE LoRA
substantially reduce reconstruction quality, and the input controls rule out
copying or a frozen-VAE round trip as explanations for the gains. Repeated
inference shows low sensitivity to the sampled Gaussian noise.

\paragraph{Efficiency Measurement Details.}
All methods are timed on the same 100 CTR-TSR-Test images at batch size 1 on
one NVIDIA RTX PRO 6000 GPU. Timing follows warm-up, excludes loading, and
synchronizes CUDA. TOLA timing includes VAE encoding, TransOCR, MoM, one IDM
evaluation, LRC, and VAE decoding; DiffTSR and TeReDiff use their full 200- and
50-step routes. MACs sum restoration-backbone evaluations at
$128{\times}512$. Counts exclude VAEs and auxiliary recognition, detection,
and spotting modules, so the reported 876.23M restoration-module count differs
from the 2.46M optimized and saved adaptation state.

\paragraph{Planned Public Release.}
Upon publication, we will release the implementation, configurations, and
exact RealCE-200 and RT50 image records under research-use terms. The release
will document the required third-party checkpoints without redistributing
their pretrained weights.

\subsection{Paired Uncertainty}

Table~\ref{tab:supp-paired-statistics} reports paired intervals, which exclude
zero for the LRC effects on PSNR, SSIM, LPIPS, and DISTS, but not NED; the exact
paired McNemar test likewise finds no significant ACC difference. Because all
comparisons use filename-aligned outputs, the intervals measure within-sample
changes and separate consistent reconstruction gains from limited recognition
changes. Paired resampling preserves image correspondence when estimating
uncertainty across methods. Predicted conditioning significantly improves ACC over
null context, whereas the GT-prior gap confirms remaining headroom from prior
errors. Figure~\ref{fig:supp-prior-controls} plots these conditioning controls,
and Figure~\ref{fig:supp-lrrb-effects} visualizes the paired LRC effects.

\par

\end{multicols}
\clearpage

{
\setlength{\parindent}{0pt}
\centering
\begin{minipage}[t]{0.55\textwidth}
\centering
\small
\setlength{\tabcolsep}{2.4pt}
\textbf{(a) Fixed-checkpoint image metrics}\par\smallskip
\begin{tabular*}{\linewidth}{@{\extracolsep{\fill}}lrrrrr@{}}
\toprule
Control & PSNR$\uparrow$ & SSIM$\uparrow$ & LPIPS$\downarrow$ & DISTS$\downarrow$ & FID$\downarrow$ \\
\midrule
Inference $t=999$ & 24.8067 & .7235 & .2289 & .1953 & 12.6849 \\
Inference $t=949$ & 24.9307 & .7257 & .2536 & .2109 & 17.8904 \\
Inference $t=899$ & 24.6970 & .7219 & .2777 & .2250 & 22.6079 \\
$\rvepsilon=\vzero$ & 23.4153 & .7014 & .2360 & .2011 & 14.7950 \\
VAE LoRA off & 23.0239 & .7188 & .2458 & .2074 & 17.4872 \\
\bottomrule
\end{tabular*}

\end{minipage}
\hfill
\begin{minipage}[t]{0.42\textwidth}
\centering
\small
\setlength{\tabcolsep}{2.4pt}
\textbf{(b) Fixed-checkpoint recognition metrics}\par\smallskip
\begin{tabular*}{\linewidth}{@{\extracolsep{\fill}}lrrrr@{}}
\toprule
Control & T-ACC$\uparrow$ & T-NED$\uparrow$ & P-ACC$\uparrow$ & P-NED$\uparrow$ \\
\midrule
Inference $t=999$ & .5391 & .7362 & .4626 & .6585 \\
Inference $t=949$ & .5402 & .7381 & .4614 & .6589 \\
Inference $t=899$ & .5395 & .7365 & .4615 & .6585 \\
$\rvepsilon=\vzero$ & .5359 & .7295 & .4539 & .6521 \\
VAE LoRA off & .5389 & .7335 & .4572 & .6552 \\
\bottomrule
\end{tabular*}

\end{minipage}

\vspace{7pt}

\begin{minipage}[c]{0.55\textwidth}
\centering
\small
\setlength{\tabcolsep}{2.4pt}
\textbf{(c) Input controls}\par\smallskip
\begin{tabular*}{\linewidth}{@{\extracolsep{\fill}}lrrrrrr@{}}
\toprule
Control & PSNR & SSIM & LPIPS & DISTS & T-ACC & T-NED \\
\midrule
LR input & 21.9863 & .5799 & .6519 & .3964 & .4235 & .6102 \\
Frozen VAE & 15.5357 & .5033 & .5913 & .4025 & .3427 & .5245 \\
\bottomrule
\end{tabular*}

\end{minipage}
\hfill
\begin{minipage}[c]{0.42\textwidth}
\centering
\small
\setlength{\tabcolsep}{3.0pt}
\textbf{(d) Repeated stochastic inference}\par\smallskip
\begin{tabular*}{\linewidth}{@{\extracolsep{\fill}}lrr@{}}
\toprule
Metric & LoRA & LoRA+LRC \\
\midrule
PSNR & $24.4242{\pm}.0019$ & $24.8082{\pm}.0013$ \\
SSIM & $.710845{\pm}.000034$ & $.723514{\pm}.000012$ \\
LPIPS & $.241043{\pm}.000078$ & $.228854{\pm}.000054$ \\
DISTS & $.197032{\pm}.000057$ & $.195314{\pm}.000031$ \\
T-ACC & $.537603{\pm}.000143$ & $.540487{\pm}.001192$ \\
T-NED & $.734259{\pm}.000461$ & $.736198{\pm}.000637$ \\
\bottomrule
\end{tabular*}

\end{minipage}
\captionof{table}{Inference controls and stochastic stability on CTR-TSR-Test
$\times4$. Panels (a)--(c) use fixed checkpoints; panel (d) reports
mean$\pm$standard deviation across repeated inference. VAE LoRA off disables
all VAE-encoder LoRA, including the latent projection, while retaining IDM
LoRA and LRC.}
\label{tab:supp-inference-stability}
\par}
\vspace{10pt}
\noindent\begin{minipage}[t]{0.48\textwidth}\vspace{0pt}

{
\setlength{\parindent}{0pt}
\centering
\small
\setlength{\tabcolsep}{1.3pt}
\begin{tabular*}{\linewidth}{@{\extracolsep{\fill}}llrr@{}}
\toprule
Comparison & Metric & Change & 95\% CI \\
\midrule
LRC--LoRA & PSNR  & $+0.3847$ & $[+0.3740,+0.3955]$ \\
          & SSIM  & $+0.0127$ & $[+0.0123,+0.0130]$ \\
          & LPIPS & $-0.0121$ & $[-0.0128,-0.0115]$ \\
          & DISTS & $-0.0017$ & $[-0.0021,-0.0013]$ \\
          & ACC ($p=0.584$) & $+0.0014$ & $[-0.0031,+0.0058]$ \\
          & NED   & $+0.0025$ & $[-0.0002,+0.0053]$ \\
\midrule
Pred.--Null & ACC ($p=0.0035$) & $+0.0083$ & $[+0.0028,+0.0138]$ \\
            & NED & $+0.0082$ & $[+0.0051,+0.0113]$ \\
GT--Pred.   & ACC ($p<0.001$) & $+0.0862$ & $[+0.0797,+0.0925]$ \\
            & NED & $+0.0716$ & $[+0.0675,+0.0757]$ \\
\bottomrule
\end{tabular*}

\captionof{table}{Filename-paired uncertainty on CTR-TSR-Test $\times4$.
Effect-size intervals use 5,000 paired bootstrap resamples; ACC significance
is additionally evaluated with the exact paired McNemar test. Changes are
computed from unrounded values.}
\label{tab:supp-paired-statistics}
\par}
\subsection{Exact Recognition Diagnostics}

Tables~\ref{tab:supp-confidence}--\ref{tab:supp-prior-strata} summarize
recognition behavior by confidence, text category, and edit operation.
Confidence denotes the mean confidence of the predicted TransOCR tokens.
Prior ACC is the exact-match accuracy of the predicted text prior, and D and T
denote DiffTSR and TOLA, respectively. The same 8,089 filename-aligned
CTR-TSR-Test $\times4$ samples and TransOCR normalization are used throughout.

The confidence relationship is monotonic, but the method improvement is not
restricted to high-confidence priors. TOLA improves ACC and NED in all four
confidence intervals and all four text categories. The edit decomposition
also shows reductions in substitutions, deletions, and insertions, lowering
the overall error rate from 42.34 to 32.55 edits per 100 characters. The large
gap between correct- and wrong-prior strata identifies prior recognition as
the primary remaining source of exact-transcription error.

\par

\par
\end{minipage}\hfill\begin{minipage}[t]{0.48\textwidth}\vspace{0pt}

{
\setlength{\parindent}{0pt}
\noindent\begin{minipage}{\linewidth}
\centering
\small
\setlength{\tabcolsep}{2.1pt}
\begin{tabular*}{\linewidth}{@{\extracolsep{\fill}}lrrrrrr@{}}
\toprule
Confidence & $N$ & Prior & D-A & T-A & D-N & T-N \\
\midrule
$[0,.25)$   & 246   & .0041 & .0407 & .1382 & .1618 & .3454 \\
$[.25,.50)$ & 1,344 & .0074 & .0476 & .1339 & .1976 & .3694 \\
$[.50,.75)$ & 1,926 & .0685 & .1506 & .2721 & .4273 & .5771 \\
$[.75,1]$   & 4,573 & .7208 & .7420 & .7923 & .9063 & .9321 \\
\bottomrule
\end{tabular*}

\captionof{table}{Exact recognition by predicted-prior confidence. D/T denote
DiffTSR/TOLA; A/N denote ACC/NED.}
\label{tab:supp-confidence}
\end{minipage}
\par}
\par\vspace{12pt}

{
\setlength{\parindent}{0pt}
\noindent\begin{minipage}{\linewidth}
\centering
\small
\setlength{\tabcolsep}{2.3pt}
\begin{tabular*}{\linewidth}{@{\extracolsep{\fill}}lrrrrr@{}}
\toprule
Category & $N$ & D-A & T-A & D-N & T-N \\
\midrule
Chinese   & 6,946 & .4587 & .5346 & .6366 & .7251 \\
English   & 884   & .5204 & .5792 & .7539 & .8098 \\
Mixed     & 120   & .1750 & .2917 & .5543 & .6435 \\
Num./Other& 139   & .6475 & .7266 & .8487 & .9038 \\
\bottomrule
\end{tabular*}

\captionof{table}{Exact recognition by text category. TOLA improves over
DiffTSR in every category.}
\label{tab:supp-script}
\end{minipage}
\par}
\par\vspace{12pt}

{
\setlength{\parindent}{0pt}
\noindent\begin{minipage}{\linewidth}
\centering
\small
\setlength{\tabcolsep}{3.2pt}
\begin{tabular*}{\linewidth}{@{\extracolsep{\fill}}lrrrrr@{}}
\toprule
Method & Sub. & Del. & Ins. & Total & Edits/100 \\
\midrule
DiffTSR & 14,657 & 3,139 & 673 & 18,469 & 42.34 \\
TOLA    & 12,082 & 1,529 & 589 & 14,200 & 32.55 \\
\bottomrule
\end{tabular*}

\captionof{table}{Character-level edits over 43,622 ground-truth characters
under the normalized TransOCR evaluation used for ACC and NED.}
\label{tab:supp-error-ops}
\end{minipage}
\par}
\par\vspace{12pt}

{
\setlength{\parindent}{0pt}
\noindent\begin{minipage}{\linewidth}
\centering
\small
\setlength{\tabcolsep}{3.0pt}
\begin{tabular*}{\linewidth}{@{\extracolsep{\fill}}lrrrr@{}}
\toprule
Prior & $N$ & Exact GT & Copy & Other \\
\midrule
Correct & 3,439 & 3,330 & -- & 109 \\
Wrong   & 4,650 & 1,031 & 376 & 3,243 \\
\bottomrule
\end{tabular*}

\captionof{table}{Behavior conditioned on predicted-prior correctness.
Counts sum to all 8,089 CTR-TSR-Test samples.}
\label{tab:supp-prior-strata}
\end{minipage}
\par}
\par\vspace{12pt}
\end{minipage}

\clearpage
\onecolumn

{
\setlength{\parindent}{0pt}
\centering
\includegraphics[width=\textwidth]{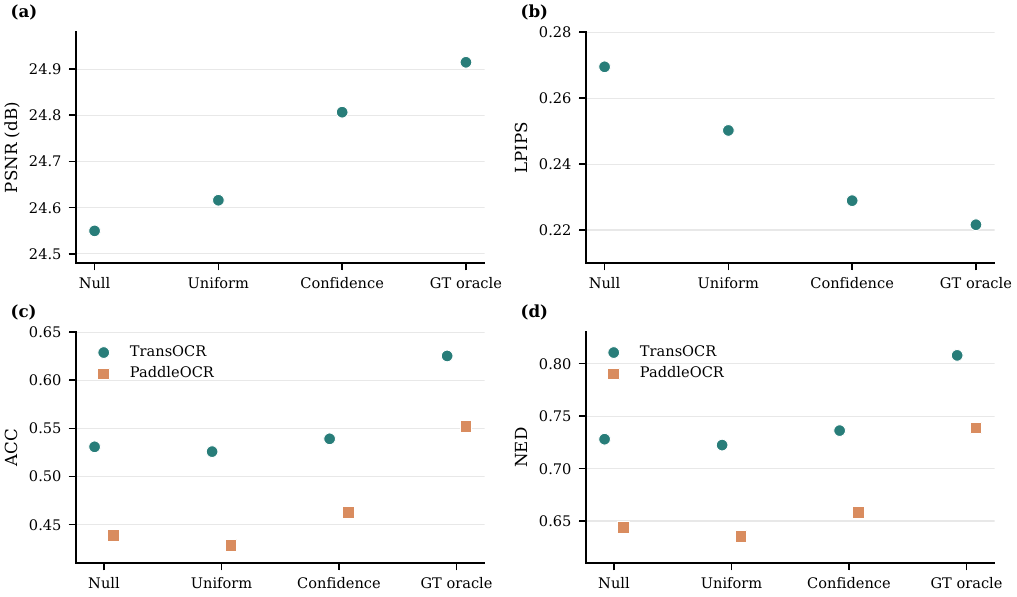}
\captionof{figure}{Text-prior controls with null (N), uniform (U), predicted (C), and
ground-truth oracle (GT) conditions.}
\label{fig:supp-prior-controls}
\par}
\vspace{12pt}
\begin{center}

{
\setlength{\parindent}{0pt}
\noindent\begin{minipage}{0.49\textwidth}
\centering
\includegraphics[width=\linewidth]{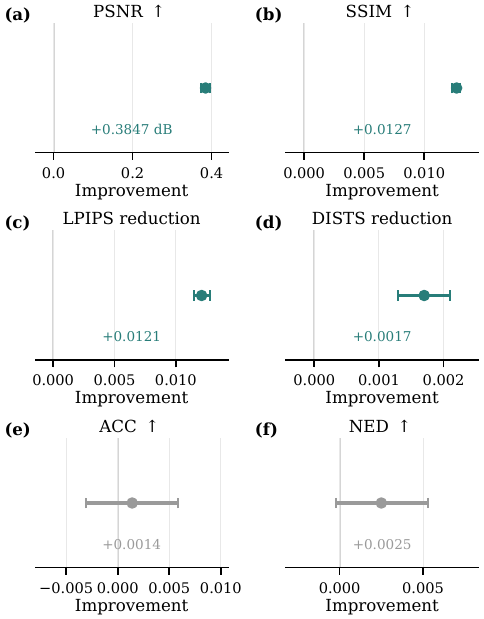}
\captionof{figure}{Paired LRC effects over LoRA-only with 95\% CIs.}
\label{fig:supp-lrrb-effects}
\end{minipage}
\par}
\end{center}